# Frontier Based Exploration for Autonomous Robot

Anirudh Topiwala
Maryland Robotics Center
University of Maryland
anirudht@umd.edu

Pranav Inani
Maryland Robotics Center
University of Maryland
inani@umd.edu

Abhishek Kathpal
Maryland Robotics Center
University of Maryland
akathpal@umd.edu

*Abstract*— **Exploration is process of selecting target points that yield the biggest contribution to a specific gain function at an initially unknown environment. Frontier-based exploration is the most common approach to exploration, wherein frontiers are regions on the boundary between open space and unexplored space. By moving to a new frontier, we can keep building the map of the environment, until there are no new frontiers left to detect. In this paper, an autonomous frontier-based exploration strategy, namely Wavefront Frontier Detector (WFD) is described and implemented on Gazebo Simulation Environment as well as on hardware platform, i.e. Kobuki TurtleBot using Robot Operating System (ROS). The advantage of this algorithm is that the robot can explore large open spaces as well as small cluttered spaces. Further, the map generated from this technique is compared and validated with the map generated using turtlebot_teleop ROS Package.**

## I. Introduction

Exploration of an unknown environment is a fundamental problem in the field of autonomous mobile robotics which deals with exploration of unknown areas while creating a map of the environment. Conventionally, human maps the environment in advance and that map is used by the robot for subsequent navigation while avoiding obstacles. Exploration has the potential to remove the human from the loop for generating a map of an unknown environment. There are many applications of exploration algorithms in areas like space robotics, sensor deployment and defense robotics etc.

In the field of autonomous exploration, a consistent and concise approach has been proposed by Brian Yamauchi in 1997 [1] which is the foundation for most of the current algorithms for exploration. The central question in exploration is: Given the present knowledge about the world, where should we move the robot to get the most information? This can be answered by the concept of Frontiers.

Frontiers are regions on the boundary between unexplored and explored space. To gain the most new and useful information, the robot must move to the frontiers and explore again.[2] By repeating this process, the mapped territory expands by pushing the boundary between the known and the unknown. When there are no new frontiers left to explore, the exploration is considered complete. There are many [7][8] other techniques based on frontiers such as Wavefront Frontier Detector (WFD) and Fast Frontier Detection [3] which reduces the time complexity of the original Frontier based exploration technique.

The focus of this paper is implementation of the Wavefront Frontier Detector Algorithm to find frontiers and then use gmapping, that implements Simultaneous Localization and Mapping (SLAM) [4]and move base [5], a planning algorithm package to plan a path to move to the nearest frontier iteratively till the complete environment is mapped. The terminology which is used in this paper [1] is discussed here:

- **Unknown Region** is the territory that has not been covered yet by the robot's sensors.
- **Known Region** is the territory that has already been covered by the robot's sensors.
- **Open-Space** is a known region which does not contain an obstacle.
- **Occupied-Space** is a known region which contains an obstacle.
- **Occupancy Grid** is a grid representation of the environment. Each cell holds a probability that represents if it is occupied.
- **Frontier** is the segment that separates known regions from unknown regions. Formally, a frontier is a set of unknown points that each have at least one open-space neighbor. The terminology described above is shown in Fig. 1.

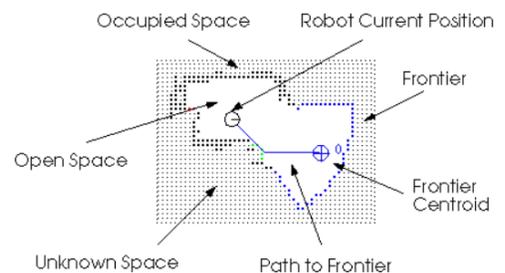

Fig. 1. Frontier

For Simulation, the algorithm is implemented on the Gazebo environment of the Robotics Realization Lab at the Maryland Robotics Center. For the hardware implementation, Turtlebot2 platform was used. The Navigation stack was used to produce a safe path for the robot to execute, by processing data from odometry [9], sensors and occupancy grid map.

The remaining paper is organized as follows. In section II, previous studies related to frontier-based techniques are discussed. Section III is focused towards autonomous map generation methodology. In Section IV, results obtained from implementation of Wavefront frontier detection technique are analyzed and compared. Section V highlights the conclusions of the paper.

## II. PREVIOUS STUDIES

The original paper by Brian Yamauchi [1] implemented the Frontier-based exploration on a Nomad 200 mobile robot equipped with a laser rangefinder, sixteen sonar sensors, and sixteen infrared sensors. Laser-limited sonar was used to build the occupancy grid, combining the data from the sonar with the data returned from the laser rangefinder.

The frontier detection algorithm is based on Breadth-First Search (BFS). First, a BFS search is run on the entire grid and frontier points are added to a queue data structure. Next, another BFS search is run on the frontier points to obtain the final frontiers. This is possible because the frontier is nothing but a connection of various frontier points. Another data structure (queue) is used which contains the centroid of the frontiers. They are arranged in the queue in the increasing order of their Euclidean distance from the current position of the Robot. So, the algorithm directs the robot to move to the closest frontier. Once the robot reaches the frontier, it performs a 360 degree sweep to create a map of that region. This process of adding frontiers and navigating to them is continued until there are no more frontiers to be explored which basically means that the entire environment has been successfully mapped.

Once frontiers have been detected within a particular evidence grid, the robot attempts to navigate to the nearest accessible, unvisited frontier. The path planner uses a breadth-first search on the grid, starting at the robot's current cell and attempting to take the shortest obstacle-free path to the cell containing the goal location. [6]

The image shown in Fig. 2 shows the path planned by the robot through a cluttered office, around a desk, and past a chair. The blue line represents the path, while the blue dots represent cells that are checked to guarantee that the space around the path is free of obstacles. The red crosshairs mark the robot's destination. The red circles indicate obstacles around which the planner had to detour the path.

While the robot moves toward its destination, reactive obstacle avoidance behaviors prevent collisions with any obstacles not present while the evidence grid was constructed. In a dynamic environment, this is necessary to avoid collisions with, for example, people who are walking about. These behaviors also allow the robot to steer around these obstacles and, as long as the world has not changed too drastically, return to follow its path to the destination [5].

When the robot reaches its destination, that location was added to the list of previously visited frontiers. The robot performs a 360-degree sensor sweep using laser-limited sonar and adds the new information to the evidence grid. Then the robot detects frontiers present in the updated grid and attempts to navigate to the nearest accessible, unvisited frontier.

If the robot is unable to make progress toward its destination, then after a certain amount of time, the robot will determine that the destination is inaccessible, and its location will be added to the list of inaccessible frontiers. Then the robot conducts a sensor sweep, updates the evidence grid, and attempts to navigate to the closest remaining accessible, unvisited frontier.

The robot is able to successfully explore and navigate within a real-world office environment. The image depicts the occupancy grids constructed, as the real robot explored an office cluttered with desks, chairs, bookshelves, cabinets, a large conference table, a sofa, and other obstacles.

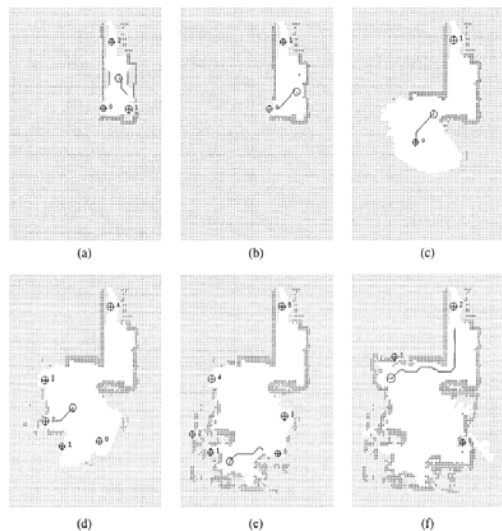
Fig. 3. Occupancy grids during exploration

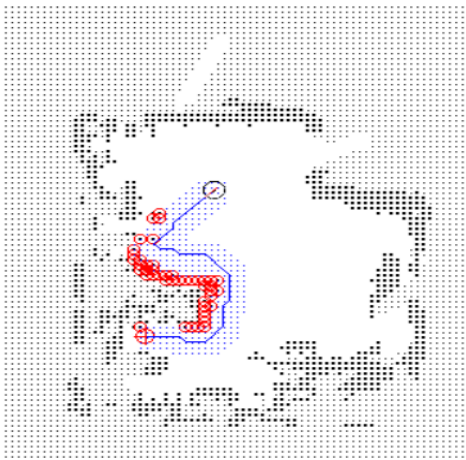
Fig. 2. Path planning for next Frontier

## III. METHODOLOGY

The algorithm implemented in this paper is an improved version of the original frontier-based exploration [1] method. It is known as the Wavefront Frontier Detector (WFD) algorithm [3]. WFD is similar to the original approach as even this algorithm is based on two nested Breadth-First Searches. The main advantage of the WFD algorithm over the original is that it only scans the known regions of the occupancy grid as opposed to the original approach which scans the entire grid at every run of the algorithm. The WFD approach is valid since frontiers never show up in the unknown region. This key difference significantly reduces the time complexity of the algorithm. To ensure that only the known regions are scanned every time the algorithm is called, WFD classifies points with one of four possible indications:

1. **Map-Open-List:** points that have been enqueued by the outer BFS.
2. **Map-Close-List:** points that have been dequeued by the outer BFS.
3. **Frontier-Open-List:** points that have been enqueued by the inner BFS.
4. **Frontier-Close-List:** points that have been dequeued by the inner BFS.

The map data structure is used to implement these four lists. In addition to avoiding rescanning of the entire occupancy grid, these lists ensure that a frontier point is not assigned to multiple frontiers. The pseudo code is shown in Fig. 4.

First, a queue data-structure is initialized to find all the frontier points in the occupancy grid. Let this be called $queue_m$. Points enqueued (resp. dequeued) to this list are marked as Map-Open-List (resp. Map-Closed-List). Only points that are not a part of Map-Open-List or Map-Close-List are considered to be added to $queue_m$. In addition to this, an additional constraint is added viz., each point being added to this queue must have at least one open-space neighbour (Line 27). This is the most crucial part of the algorithm as it ensures that only known regions are scanned.

At the start, the point in the occupancy grid corresponding to the current pose of the robot is enqueued into a queue data-structure. Next, a BFS is performed till a frontier point is encountered. Once, such a point is encountered, a new queue is initialized to extract its frontier. Let this be called $queue_f$. Only points that are not a part of the Map-Open-List are enqueued to this queue (Line 6). The frontier point obtained from the outer BFS is first enqueued to this queue. Now, a fresh BFS is performed to find all connected frontier points to extract its frontier.

All points being visited in the inner BFS are marked as Frontier-Open-List. Whenever a frontier point is encountered in this BFS, it is stored in a list containing all other frontier points in connection with their respective frontier. These points are marked as Frontier-Closed-List. A frontier point is added to a frontier list only if it is not a part of the Frontier-Open-List, Frontier-Close-List or Map-Closed-List (Line 20). This ensures that each frontier point is assigned to a unique frontier. Using this scanning policy, it is guaranteed that all frontier points will be assigned to a unique frontier at the end of each algorithm run.

---

**Require:** $queue_m$ // queue, used for detecting frontier points from a given map
**Require:** $queue_f$ // queue, used for extracting a frontier from a given frontier cell
**Require:** $pose$ // current global position of the robot

1: $queue_m \leftarrow \emptyset$
2: ENQUEUE($queue_m$, $pose$)
3: mark $pose$ as "Map-Open-List"
4: **while** $queue_m$ is not empty **do**
5:    $p \leftarrow$ DEQUEUE($queue_m$)
6:    **if** $p$ is marked as "Map-Close-List" **then**
7:      continue
8:    **if** $p$ is a frontier point **then**
9:      $queue_f \leftarrow \emptyset$
10:     $NewFrontier \leftarrow \emptyset$
11:     ENQUEUE($queue_f$, $p$)
12:     mark $p$ as "Frontier-Open-List"
13:     **while** $queue_f$ is not empty **do**
14:       $q \leftarrow$ DEQUEUE($queue_f$)
15:       **if** $q$ is marked as {"Map-Close-List","Frontier-Close-List"} **then**
16:         continue
17:       **if** $q$ is a frontier point **then**
18:         add $q$ to $NewFrontier$
19:         **for all** $w \in adj(q)$ **do**
20:           **if** $w$ not marked as {"Frontier-Open-List","Frontier-Close-List", "Map-Close-List"} **then**
21:             ENQUEUE($queue_f$,$w$)
22:             mark $w$ as "Frontier-Open-List"
23:       mark $q$ as "Frontier-Close-List"
24:     save data of $NewFrontier$
25:     mark all points of $NewFrontier$ as "Map-Close-List"
26:    **for all** $v \in adj(p)$ **do**
27:      **if** $v$ not marked as {"Map-Open-List","Map-Close-List"} and $v$ has at least one "Map-Open-Space" neighbor **then**
28:         ENQUEUE($queue_m$,$v$)
29:         mark $v$ as "Map-Open-List"
30:    mark $p$ as "Map-Close-List"

Fig. 4. Pseudo-code of WFD algorithm

This algorithm is incorporated with ROS [10] to navigate and map an unknown environment using a TurtleBot2. The robot is spawned in a completely unknown environment. The gmapping node is running in the background to map anything the robot's sensor picks up. Initially, the robot is programmed to rotate to get the nearby environment.

Next, the ROS node running the WFD algorithm extracts all the frontiers in this partially known map. Now, the median of each of these frontiers are calculated and arranged in the increasing order of the Euclidean distance from the current robot pose. Now using the move_base package of ROS, the robot is made to move to the closest frontier median. In the case that move_base is unable to plan a path to the nearest frontier median; the next closest frontier is chosen and so on. Once the median point is reached, the robot performs a 360-degree rotation in order to scan the nearby environment.

It should be noted that the choice of choosing the median point instead of the centroid is a slight departure from the original algorithm as well as WFD. This choice was made as sometimes, the centroid may lie deep inside an unknown region, which may possibly be an obstacle. In such a scenario, move_base will fail to plan a path. By moving to the median point, we ensure exploration as well as guaranteed path planning.

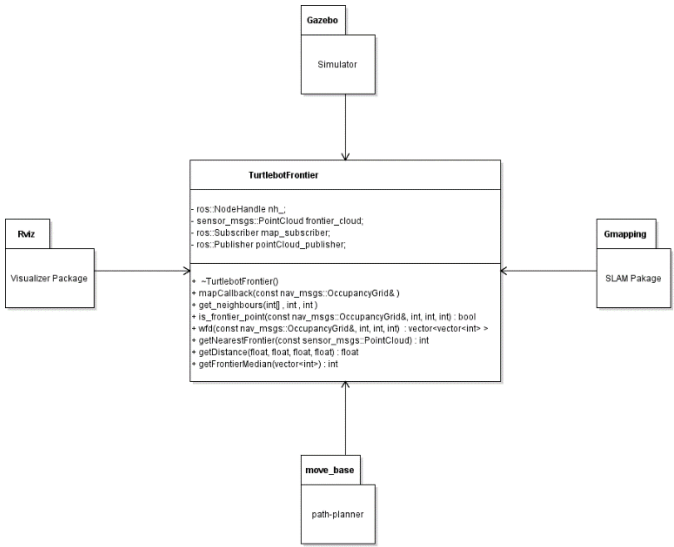

Fig 5: Software Framework

The algorithm is called iteratively, and the robot is moved to the frontier medians until there are no more frontiers left in the occupancy grid. Once such a state is reached it indicates that the given map was successfully mapped. The map can then be saved using the map_server node of ROS.

## IV. RESULTS

The WFD exploration algorithm was implemented in both simulated as well as real environment. In order to validate the results of the WFD algorithm, map generated by ROS turtlebot_teleop package is used. The implementation is done using ROS framework. Gazebo, rviz, gmapping and move_base ROS packages are used for generating the map. The experimental results are divided in the following three sections – Simulation Results, Hardware Implementation and comparison with manually generated map.

### A. Simulation Results

The simulated environment used for autonomous navigation is a replication of Robotics Realization Lab (RRL) of Maryland Robotics Centre. The top view of the environment is shown in Fig. 6. which is showing the robot in Gazebo and Rviz. As we can see in Rviz, the robot has no information of the surrounding data and therefore, the first step here is to rotate the robot.

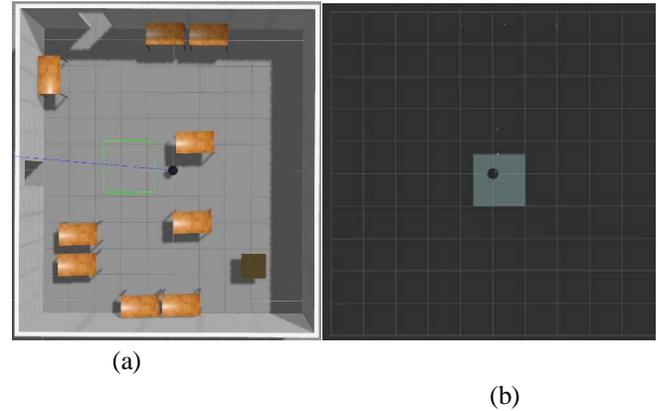

(a)

(b)

Fig. 6. (a) Gazebo environment (RRL lab) (b)Rviz Visualization

Once the robot is rotated, we will have a set of map data which the WFD algorithm will process to return the frontier point. Fig. 7. (a) shows the map data formed in Rviz and Fig. 7. (b) is a 2D map generated representing the Frontier Points (Green region), Open space (Blue region) and Obstacles (Red). Based on this data, we will get the median frontier point which is then passed to move base as a goal point.

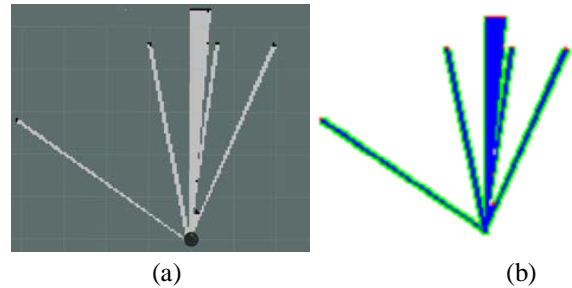

(a) (b)

Fig 7. (a) Map after Initial Rotation (b) Frontier Points Detected

Move Base will use local planner (DWA planner) and global planner (navfn) configured with the correct local and global parameters to plan a path to the goal position. Fig. 8. shows the trajectory of the path found showing the robot moving to the next frontier.

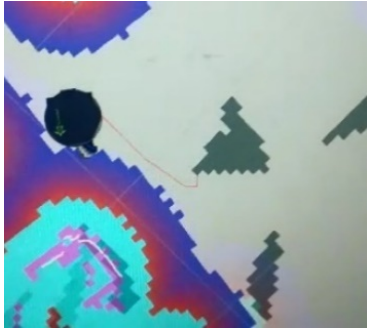

Fig. 8. Move Base Trajectory

While the goal position is being achieved, gmapping is simultaneously running in backend which would have added much more new data to the map and therefore new frontiers would be found. This process is repeated iteratively until there are no new frontiers left for the robot to go to. Sometimes even though there is enough gmapping data available, the planner is not able to find a valid trajectory. In such times, a recovery behaviour of rotating the robot 360-degress is activated. This will help gather more surrounding data and get a proper trajectory plan.

The following figures show the expansion of the map generated by the turtlebot2.

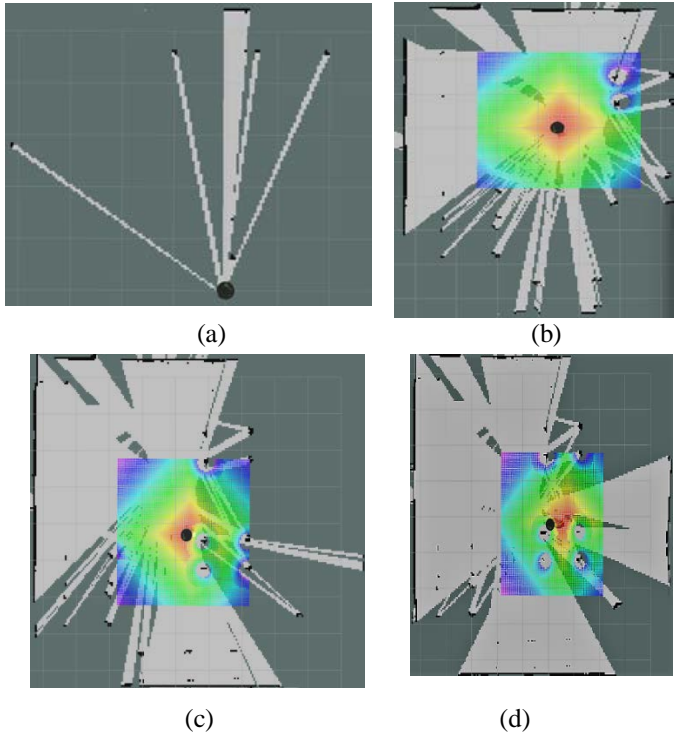

(a) (b)

(c) (d)

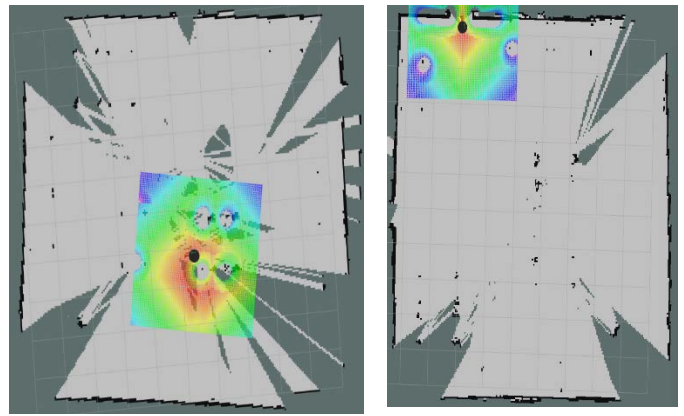

(e) (f)

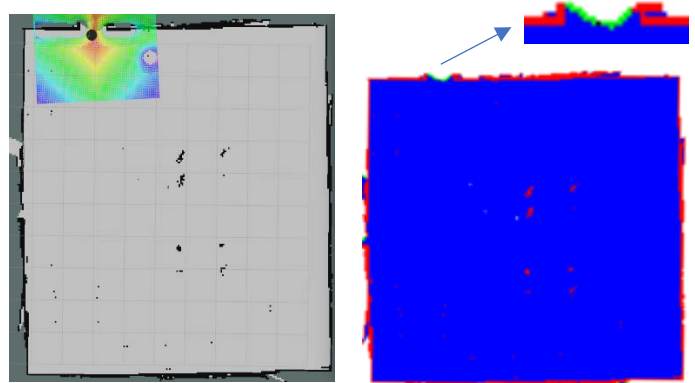

(g) (h)

Fig. 9. (a -g) Represents the gradual exploration of the RRL lab by turtlebot2. (h) Gmapping data Representation.

It can be seen in Fig. 8 that the turtleBot starts with minimal data and from the frontiers generated, it navigates to new frontiers. The final map generated by the turtleBot is shown in Fig. 9. (g). The map generated is identical to the one shown in gazebo in Fig. 6(a). The black dots seen in the map are the obstacles or the table legs as seen in gazebo. Fig. 9(h) depicts the final map generated from the gmapping data obtained. We can see that there no new frontiers in the given scene. The only frontiers exist are at the door of the scene which leads to open space and that is where the robot is trying to go. Therefore, using Wavefront Frontier Detector algorithm the turtleBot2 can map the entirely unknown environment.

### B. Hardware Implementation

For hardware implementation, TurtleBot2 platform is used. The sensor data is taken from Orbbec Astra Pro which is the default sensor for TurtleBot 2. Fig 10 shows the major hardware components of TurtleBot 2.

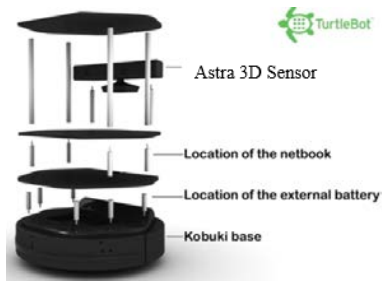

Fig. 10. Turtlebot 2 (Koubuki Base)

Navfn is used as global planner, which uses A* or Dijkstra algorithm to find a global path with minimum cost between start point and end point. The local planner used is Dynamic Window Approach (DWA) planner [11]. The goal of DWA is to produce a (v, ω) pair which represents a circular trajectory that is optimal for robot's local condition in a particular window. DWA maximizes an objective function that depends on (1) the progress to the target, (2) clearance from obstacles, and (3) forward velocity to produce the optimal velocity pair. The clearance from obstacle or padding is decided by cost map provided. TurtleBot inbuilt parameters are used to define the local and the global cost map and tuned it to get the best results. Thus, once all the parameters are set, optimal path is computed between initial and goal point. The final explored map by the TurtleBot 2 robot is shown in Fig 11. The map was generated in approximately 17 minutes.

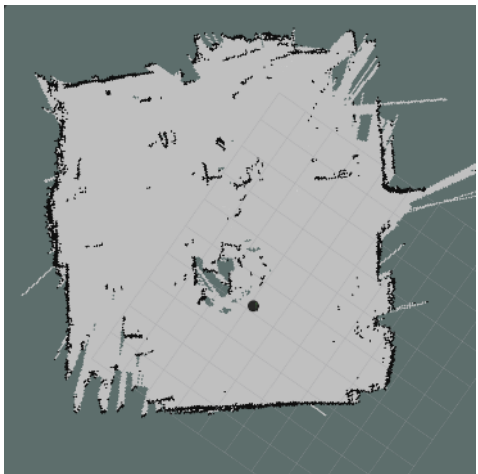

Fig. 11. Explored map by actual Turtlebot

The difference between hardware output and simulation is because of several reasons. Firstly, the gazebo environment is not an exact replica of the actual world. Also, during exploration, the DWA local planner continuously performs object avoidance to avoid dynamic obstacles, which is never the case in simulation. Also, the parameter tuning for the planner is different for hardware and the simulation. Therefore, getting accurate results require a further fine tuning of the planner on TurtleBot2 [5].

*C. Comparison with manually generated map*

The map generated from the autonomous exploration is validated and compared by the manually map generated by using "turtlebot_teleop" and "gmapping" package. The turtlebot_teleop package provides the launch files to move the TurtleBot using keyboard. Gmapping package is used to update the map as the TurtleBot move in the real or simulated environment. The map of the simulated RRL environment generated from the turtlebot_teleop package along with the WFD output is shown in Fig .12

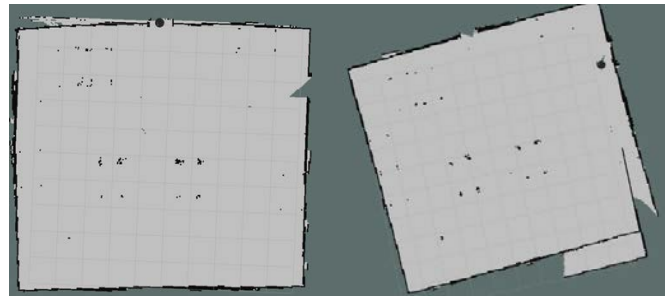

(a)                 (b)

Fig. 12. (a) WFD Output (b) Teleoperated Output

As it can be observed in Fig.9 that the map generated from WFD is more accurate. The output from turtlebot_teleop package varies from user to user. Depending on the maximum velocity provided as input by user, the map boundaries may not be accurate.

V. CONCLUSION

In this paper, Wavefront Frontier Detector algorithm for autonomous exploration is described and implemented on TurtleBot2 Kobuki platform. Different simulation and real environment results are generated. For the simulation purposes, Gazebo and Rviz are used for proper visualization. It is concluded that the output generated by autonomous exploration gives better results rather than having manually generated map using turtlebot_teleop package. Obstacle avoidance is also carried out with the help of local and global cost maps. Local cost map is generated using Astra camera sensor on turtlebot2. Global cost maps are generated using the mapping algorithm.

The output generated from WFD gives much smoother trajectories and requires less time and effort in comparison to manually generating the map. It has also been observed that total time of travel and total distance travelled to map the entire area significantly reduced when using WFD algorithm in comparison to a human tele-operating the robot to generate the map.